\documentclass{article}
\usepackage{spconf,amsmath,graphicx}
\usepackage{multirow}
\usepackage{makecell}
\usepackage{colortbl}
\usepackage[table,xcdraw,dvipsnames]{xcolor}
\usepackage{algorithm}
\usepackage{algpseudocode}  
\usepackage{fdsymbol}
\usepackage[misc,geometry]{ifsym}
\usepackage{xspace}
\usepackage{hyperref}
\usepackage{arydshln}
\usepackage{enumitem}

\usepackage{atbegshi,picture}\AtBeginShipoutNext{\AtBeginShipoutUpperLeft{\scriptsize \put(0.9cm,-.02\paperheight){\parbox{1.1\textwidth}{\textcolor[gray]{0.5}{\copyright~2022~IEEE. Personal use of this material is permitted. Permission from IEEE must be obtained for all other uses, in any current or future media, including reprinting/republishing this material for advertising or promotional purposes, creating new collective works, for resale or redistribution to servers or lists, or reuse of any copyrighted component of this work in other works.}}}}}

\newcommand{\OURS}{\textsf{GDAP}\xspace}

\usepackage{cleveref}
\crefname{section}{§}{§§}

\title{Generating Disentangled Arguments with Prompts: \\A Simple Event Extraction Framework that Works}
\name{Jinghui Si\textsuperscript{* $\clubsuit\spadesuit\heartsuit$}\thanks{* indicates equal contribution. \text{\Letter} indicates the corresponding author.} \quad Xutan Peng\textsuperscript{* $\diamondsuit$} \quad Chen Li\textsuperscript{ $\clubsuit\spadesuit$} \quad Haotian Xu\textsuperscript{ $\heartsuit$} \quad Jianxin Li\textsuperscript{ $\clubsuit\spadesuit$ \text{\Letter}}}
\address{\textsuperscript{$\clubsuit$}Beijing Advanced Innovation Center for Big Data and Brain Computing, Beihang University, China
\\\textsuperscript{$\spadesuit$}State Key Laboratory of Software Development Environment, Beihang University, China
\\\textsuperscript{$\heartsuit$}Alibaba Group, China \quad  \textsuperscript{$\diamondsuit$}Department of Computer Science, The University of Sheffield, UK
\\ \normalsize \{sijh, lichen, lijx\}@act.buaa.edu.cn \quad x.peng@shef.ac.uk \quad albert.xht@alibaba-inc.com \quad
}
\definecolor{ddg}{rgb}{0.48,0.48,0.48}
\definecolor{dg}{rgb}{0.58,0.58,0.58}
\definecolor{g}{rgb}{0.68,0.68,0.68}
\definecolor{lg}{rgb}{0.78,0.78,0.78}
\definecolor{llg}{rgb}{0.9,0.9,0.9}

\begin{document}
%
\maketitle
\begin{abstract}
Event Extraction bridges the gap between text and event signals. Based on the assumption of trigger-argument dependency, existing approaches have achieved state-of-the-art performance with expert-designed templates or complicated decoding constraints. In this paper, for the first time we introduce the prompt-based learning strategy to the domain of Event Extraction, which empowers the automatic exploitation of label semantics on both input and output sides. To validate the effectiveness of the proposed generative method, we conduct extensive experiments with 11 diverse baselines. Empirical results show that, in terms of F1 score on Argument Extraction, our simple architecture is stronger than any other generative counterpart and even competitive with algorithms that require template engineering. Regarding the measure of recall, it sets new overall records for both Argument and Trigger Extractions. We hereby recommend this framework to the community, with the code publicly available at \url{https://github.com/RingBDStack/GDAP}.
\end{abstract}


\section{Introduction}
\label{sec:intro}

\begin{figure}[tpb]
    \centering
    \includegraphics[width=\linewidth]{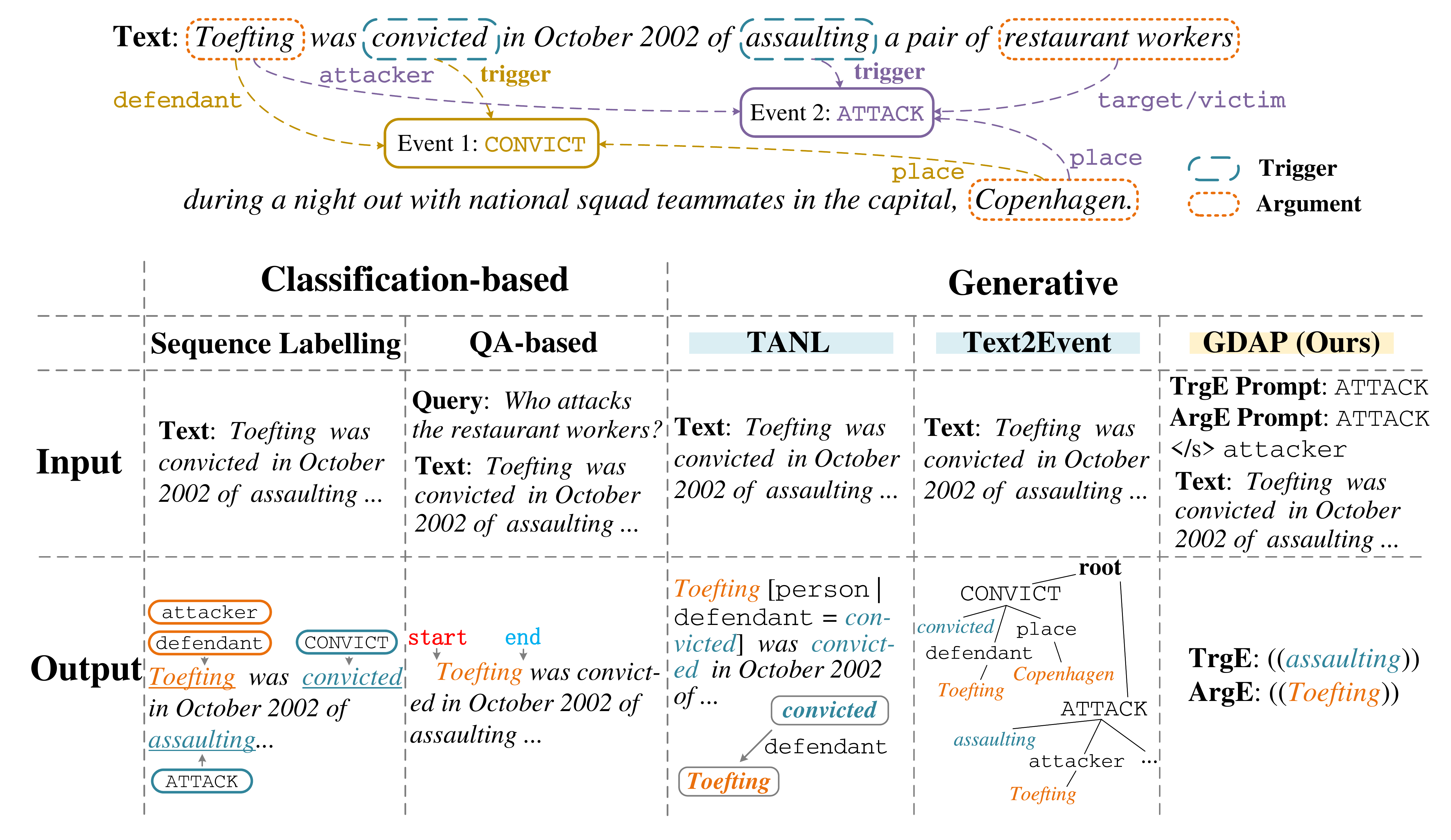}
    \caption{
        Upper: A sentence with annotated event records. Lower: The taxonomy of Event Extraction algorithms. We highlight TANL and Text2Event here, as they are most relevant to the proposed method.
        }
    \label{fig:taxomomy}
\end{figure}

Event Extraction, which aims to extract structured event signals from plain text, is a crucial but challenging Information Extraction task~\cite{ED-icassp2021,ace2005-annotation,EEsurvey-2020,HaoPengWWW,haopeng_tpami}. In the literature, an event is typically defined by a schema, which includes the event type and a set of corresponding roles. Generally speaking, to fill in this schema, an Event Extraction system needs to find \textit{triggers} that suggest an event, and ultimately, to locate the \textit{arguments} that play different roles. 
Fig.~\ref{fig:taxomomy} illustrates a real-world example with two events. For the event `\texttt{CONVICT}' which can be triggered by `\textit{convicted}', we need to extract arguments `\textit{Toefting}' and `\textit{Copenhagen}' for roles `\texttt{defendant}' and `\texttt{place}'. As for another event `\texttt{ATTACK}' where `\textit{assaulting}' serves as a trigger, its linked arguments `\textit{Toefting}', `\textit{Copenhagen}', and `\textit{restaurant workers}' should be picked for roles `\texttt{attacker}', `\texttt{place}', and `\texttt{target}/\texttt{victim}', respectively.
 
\begin{figure*}[ht]
    \centering
    \includegraphics[width=0.85\textwidth]{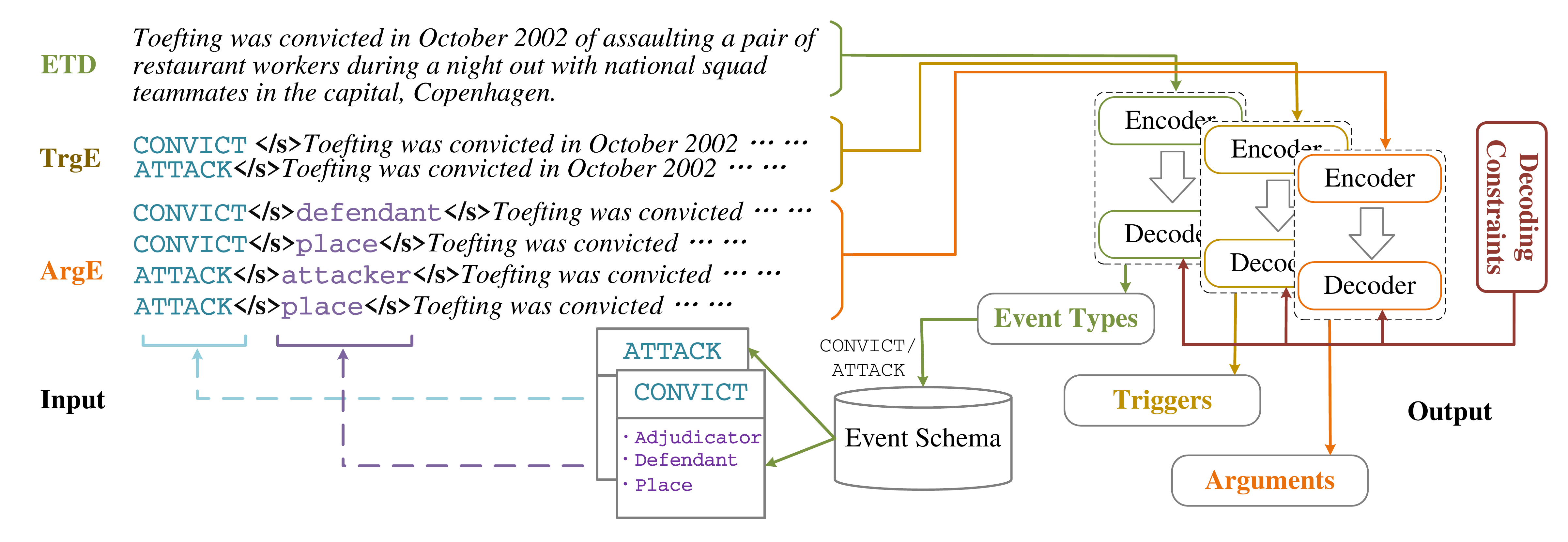}
    \caption{
        Architecture overview. \textbf{ETD}, \textbf{TrgE}, and \textbf{ArgE} are the abbreviation of Event Type Detection, Trigger Extraction, and Argument Extraction, respectively. For output examples of TrgE and ArgE, please see the lower right corner of Fig.~\ref{fig:taxomomy}.
        }
    \label{fig:overview}
\end{figure*}%

Early studies formulate Event Extraction as a token-level classification problem, i.e., to directly locate the triggers and arguments in the text and identify their categories. Many of these works simply adopt sequence labelling techniques based on Neural Networks~\cite{chen-etal-2015-event,JRNN-2016,dbRNN-2018-aaai,liu-etal-2018-jointly,PLMEE2019,OneIE-2020}. However, such methods only capture the internal pattern of input sequences without utilising the knowledge of label semantics. Therefore, another research strand, namely QA-based approaches, emerges~\cite{liu-RCEE-2020,du-cardie-2020-event,li-etal-2020-event}. With prepared templates, they first augment the training corpus 
by generating questions that are respectively targeting event types, triggers, and arguments. Next, the models learn to locate spans in the original sentences as answers, thus explicitly introducing the label knowledge.  Nevertheless, the performance of these methods heavily depends on the quality of question templates, while designing them requires high-level expertise and massive human labour.

Very recently, instead of following the classification paradigm, a new wave of algorithms frame Event Extraction as a generation task. TANL~\cite{TANL}, which is a pipeline powered by the pre-trained T5~\cite{JMLR2020-T5}, learns to \textit{sequentially} `translate' plain text input into sequences where event triggers and arguments are marked, respectively. Another T5-based generative model, namely Text2Event~\cite{lu2021text2event}, instead attempts an end-to-end manner where the output has a complicated tree-based structure. Thanks to the success of large-scale pre-trained language models, such generative approaches can reduce the manual engineering for the templates to the minimum, hence superior to the  aforementioned QA-based  methods. Nevertheless, they still exhibit bottlenecks that limit real-world applications. (1)~They focus on incorporating the label semantics (as constraints) during decoding but fail to fully exploit such signals (e.g., event types and triggers) on the encoding side.  (2)~Akin to their classification-based counterparts, generative models assume dependency between Trigger and Argument  Extractions, thus implementing these two modules either serially or jointly. However, this long-standing hypothesis is challenged by our observations, e.g., in a simple input sentence `\textit{We put the Shah of Iran in power}', the trigger `\textit{put}' is hardly beneficial to extracting arguments `\textit{Shah}' and `\textit{Iran}' either grammar or semantically. Worse still, recent findings show reveal the existence of \textit{overlapped} triggers and
arguments~\cite{CasEE}, which even leads to precision drop. (3)~For TANL, a considerable percentage of the generated tokens are task-irrelevant; for Text2Event, the output structure can be too complex to scale.

To alleviate the above issues, in this paper, we propose a novel framework that \textbf{g}enerates \textbf{d}isentangled \textbf{a}rguments with \textbf{p}rompts (\OURS). As its name suggests, \OURS achieves three remarkable algorithmic improvements. (1)~To effectively inject knowledge via various label semantics when encoding the input, for the first time we introduce prompt-based learning to the domain of Event Extraction. (2)~Unlike \textit{all} existing methods, \OURS disentangles the extraction of triggers and arguments, substantially enhancing the computational parallelism and naturally tackling the issue of overlapping. (3)~With both the architecture and the output format hugely simplified, \OURS is easy to be implemented and extended. To empirically verify the effectiveness of these advancements, we conduct extensive experiments on the standard ACE 2005 benchmark~\cite{ace2005-annotation}, where 11 strong baselines (both classification-based and generative) are involved. In the Argument Extraction task, \OURS yields the best F1 score among all generative methods, which is even competitive with state-of-the-art baselines that rely on hand-designed templates. Moreover, \OURS scores the overall highest recall in both Argument and Trigger Extractions, indicating promising applications in commercial scenarios.

\section{Method}\label{sec:methdology}

As shown in Fig.~\ref{fig:overview}, \OURS possesses three functional modules, namely Event Type Detection, Trigger Extraction, and Argument Extraction. In practice, the high diversity of event types in the schema will lead to a large range of potential trigger and argument selections, making a comprehensive traversal too expensive to afford. Thereupon, all input sentences will first pass through the Event Type Detection module to reduce computational overhead. Based on the predicted event types, \OURS will then process Trigger and Argument Extractions \textit{independently} and \textit{simultaneously}. As discussed in \cref{sec:intro}, this is the first attempt of applying such a disentangled design on Event Extraction, to our knowledge. For simplicity, all these three modules hold a similar architecture while being independently trained without parameter sharing, i.e., an encoder-decoder network based on a pre-trained language model. Please refer to details in the subsequent paragraphs.

\noindent\textbf{\underline{Event Type Detection.}} This module learns to encode a raw sentence and decode its $x$ event types using the Parenthesis Representation~\cite{para-rep} as
\\\centerline{$\mathrm{((}\mathbf{ET}_1\mathrm{)}\mathrm{(}\mathbf{ET}_2\mathrm{)} \cdots \mathrm{(}\mathbf{ET}_i\mathrm{)} \cdots \mathrm{(}\mathbf{ET}_x\mathrm{))}$,}
where $\mathbf{ET}_i$ denotes the $i$-th event type and is enclosed by special symbols `$\mathrm{(}$' and `$\mathrm{)}$', e.g., the golden output for the sentence in Fig.~\ref{fig:taxomomy} is `$\mathrm{((\texttt{CONVICT})(\texttt{ATTACK}))}$'. Due to the constraints of this special output format, the conventional decoding algorithms for text generation (e.g., greedy search and beam search), which step-wisely select the token purely based on prediction probability, cannot warrant structural validity here. Inspired by Text2Event, we design a finite-state machine whose states of token production (whether to decode `$\mathrm{(}$', `$\mathrm{)}$', or a event type) is determined by the counts of already generated `$\mathrm{(}$' and `$\mathrm{)}$'. Besides, when decoding event types, the subword vocabulary may form tokens that are not within the candidate pool, e.g., `\textit{TCONTTVIC}' is a false generation using subwords `\textit{CON}', `\textit{VIC}', and `\textit{T}'. Therefore, we turn to the tire-based constraint decoding algorithm~\cite{chen-etal-2020-parallel, decao2020autoregressive}, which guarantees the token validness by ensuring the search is only performed within a pre-built subword tree.

\noindent\textbf{\underline{Trigger Extraction.}} We introduce the recipe of prompt-based learning to this module. The input is composed of a sentence $\mathbf{Sent}$, one already detected event type $\mathbf{ET}_i$, and a special separating token (denoted as $\mathcal{T}_{sep}$; in practice we implement it as `$<$/s$>$', see \cref{ssec:exp_setup}), as \\\centerline{$\mathbf{ET}_i~\mathcal{T}_{sep}~\mathbf{Sent}$.} 
While previous methods either fail to integrate label semantics during decoding or can only import such information through templates designed by experts, we find that the very simple prompt can effectively instruct \OURS to extract triggers relevant to the semantics of the event type label, in a fully data-driven fashion. Concretely speaking, if $\mathbf{Sent}$ contains $y$ triggers corresponding to $\mathbf{ET}_i$, the expected output is 
\\\centerline{$\mathrm{((}\mathbf{Trg}_1\mathrm{)}\mathrm{(}\mathbf{Trg}_2\mathrm{)} \cdots  \mathrm{(}\mathbf{Trg}_y\mathrm{))}$,}
where $\mathbf{Trg}_1$ to $\mathbf{Trg}_y$ all come from the vocabulary of $\mathbf{Sent}$. As the format here is similar to that of the Event Type Detection module, at the decoding stage we adopt the same mechanism, i.e., the aforementioned  finite-state machine and the tire-based constraint decoding algorithm.

\noindent\textbf{\underline{Argument Extraction.}} Like the Trigger Extraction module, our Argument Extraction module also attends to the composition of a prompt and the input sentence:
\\\centerline{$\mathbf{ET}_i~\mathcal{T}_{sep}~\mathbf{RT}_{ij}~\mathcal{T}_{sep}~\mathbf{Sent}$,} 
where $\mathbf{RT}_{ij}$ is the $j$-th role type relevant to $\mathbf{ET}_{i}$ and can be decided by querying the established event schemes, e.g., role types of event `\texttt{CONVICT}' are `\texttt{defendant}' and `\texttt{place}' (see \cref{sec:intro}). As for the decoder side, if $z$ arguments are obtained, the Argument Extraction module outputs a sequence with format similar to that of the Trigger Extraction module, as
\\\centerline{$\mathrm{((}\mathbf{Arg}_1\mathrm{)}\mathrm{(}\mathbf{Arg}_2\mathrm{)} \cdots  \mathrm{(}\mathbf{Arg}_z\mathrm{))}$,} 
where $\mathbf{Arg}_1$ to $\mathbf{Arg}_z$ are also from the vocabulary of $\mathbf{Sent}$. 

We argue that apart from the enhanced encoder that can absorb valuable label semantics, the decoder of \OURS also achieves outstanding advancements beyond existing generative Event Extraction algorithms. On the one hand, although a large partition of words in $\mathbf{Sent}$ are irrelevant to Event Extraction, they are still included by TANL. In contrast, the output of \OURS only contains the extracted targets (triggers or arguments) without redundancy, which significantly improves data efficiency. On the other hand, while the tree-based decoding format of Text2Event is very complex and thus hard to scale, the generating format of \OURS is a simple list-style sequence and can therefore be easily extended to other tasks. We leave exploring this direction as an important future work.

\noindent\textbf{\underline{Negative Sampling.}}
When training the modules for Trigger and Argument Extractions, we introduce a simple yet effective negative sampling mechanism that makes our model more fault-tolerant. To be exact, for each $\mathbf{Sent}$, we randomly select $N$ event types that have not appeared. The model should learn \textit{not} to extract triggers or arguments when such negative samples appear in the prompt; instead, it should only generate an empty sequence, i.e., `$\mathrm{(())}$'.  It is worth noting that while increasing $N$ contributes to the extraction robustness, it can lead to a significant training time boost as the number of training samples grows by approximately $N+1$ times.

\section{Experiments}\label{sec:exp}

\subsection{Setup}\label{ssec:exp_setup}

\noindent\textbf{\underline{Dataset.}} The English partition in the ACE 2005 benchmark~\cite{ace2005-annotation} is the \textit{de facto} standard of Event Extraction tests. It has 599 documents annotated by 33 different event types. We adopt the popular splits released by \cite{wadden-etal-2019-entity}, where there are respectively 17172, 923, and 832 sentences for training, validating, and testing. We also perform the preprocessing steps using the script of \cite{wadden-etal-2019-entity}.

\noindent\textbf{\underline{Baselines.}} 
To evaluate the Event Extraction efficacy of GDAP, we consider 11 strong baselines from a wide range, including (1) \emph{methods based on sequence labelling}: LSTM-based \textbf{dbRNN}~\cite{dbRNN-2018-aaai}, RNN/GCN-based \textbf{JMEE}~\cite{liu-etal-2018-jointly}, BiGRU-based \textbf{Joint3EE}~\cite{Nguyen-Nguyen:2019:AAAI2019}, BERT-based \textbf{DYGIE++}~\cite{wadden-etal-2019-entity}, ELMo-based \textbf{GAIL}~\cite{Zhang:2019:GAIL}; (2) \emph{QA-based methods}: 
element-centred \textbf{BERT\_QA}~\cite{du-cardie-2020-event}, multi-turn \textbf{MQAEE}~\cite{li-etal-2020-event}, style-transfer-inspired \textbf{RCEE\_ER}~\cite{liu-RCEE-2020}; (3) \emph{generative methods}: \textbf{TANL}~\cite{TANL} and  \textbf{Text2Event}~\cite{lu2021text2event} (recall \cref{sec:intro} for detailed introductions). For fair comparisons, all baselines (including JMEE and RCEE\_ER) and our method do not utilise golden entities as they are unlikely to be available in real-world settings.

\noindent\textbf{\underline{Configurations of \OURS.}} Parallel to the generative baselines (TANL and Text2Event), \OURS adopts the pre-trained T5 as the backbone for each module, with both base (\textbf{T5-B}) and large (\textbf{T5-L}) versions tested. To align with the original implementation of T5, we choose `$<$/s$>$' as the separating token $\mathcal{T}_{sep}$. We leverage golden event type labels when composing prompts during training. Throughout all experiments, for cost-performance tradeoff, we set $N$ in negative sampling at 4 and 2 for Trigger and Argument Extractions, respectively. Identical to Text2Event, we fix the random seed at 421. The learning rate is set at 5e-5. We utilise label smoothing~\cite{MullerKH19} and AdamW~\cite{loshchilov2018decoupled}, and try the number of epochs within $\{20, 25, 30\}$ to optimise validating scores.

\noindent\textbf{\underline{Metrics.}} 
Following past studies~\cite{chen-etal-2015-event,liu-etal-2018-jointly},  we report the precision~(\textbf{P}), recall~(\textbf{R}), and F1 score (\textbf{F1}) of Trigger and Argument Extractions. Note that the output is marked as correct only when both text spans and predicted labels match with the ground-truth reference. In most industrial scenarios, arguments are the end product of an event extraction system, hence we attach greater importance to Argument Extraction than Trigger Extraction in this paper. 

\begin{table}[t]
	\small
	\centering
	\caption{
		Results of Event Extraction tests. Baseline performance is adapted from the original publications (NB: TANL has not attempted T5-L). \textbf{Bold} and \underline{underlined} numbers are the best results for all and generative models, respectively.  
	}
	\label{tab:mainresult}
	\begin{tabular}{lcccccc}		
		\hline
         & \multicolumn{3}{c}{\textbf{Trigger}} & \multicolumn{3}{c}{\textbf{Argument $\largewhitestar$}} \\
        \cline{2-4} \cline{5-7} (\%) & \textbf{P} & \textbf{R} & \textbf{F1} & \textbf{P} & \textbf{R} &  \textbf{F1} \\
 		\hline
 		\multicolumn{7}{l}{\textit{Classification-based}} \\
 		\hline
		dbRNN  & - & - & 69.6 & - & - & 50.1\\
		JMEE & - & - & - & - & - & 50.4\\
		Joint3EE & - & - & 69.8 & 52.1 & 52.1 & 52.1\\
		DYGIE++ & - & - & 69.7 & - & - & 48.8\\
		GAIL & \textbf{74.8} & 69.4 & 72.0 & \textbf{61.6} & 45.7 & 52.4\\
		BERT\_QA & 71.1 & 73.7 & \textbf{72.4} & 56.8 & 50.2 & 53.3\\
		MQAEE & - & - & 71.7 & - & - & 53.4\\
		RCEE\_ER  & - & - & -& - & - & \textbf{58.7}\\
 		\hline
 		\multicolumn{7}{l}{\textit{Generative}} \\
 		\hline
		TANL (T5-B)& - & - & 68.4 & - & - & 47.6\\
		Text2Event (T5-B) & 67.5 & 71.2 & 69.2 & 46.7 & 53.4 & 49.8\\
		Text2Event (T5-L) & \underline{69.6} & 74.4 & \underline{71.9} & \underline{52.5} & 55.2 & 53.8\\
		\hdashline
		\OURS (T5-B) & 66.1 & \underline{\textbf{75.3}} & 70.4 & \cellcolor{BlueGreen!15}{47.3} & \cellcolor{BlueGreen!15}{59.1} & \cellcolor{BlueGreen!15}{52.6}\\
		\OURS (T5-L) & 65.6 & 74.7 & 69.9 & \cellcolor{BlueGreen!15}{48.0} & \underline{\textbf{\cellcolor{BlueGreen!15}{61.6}}} & \underline{\cellcolor{BlueGreen!15}{54.0}}\\
		\hline
	\end{tabular}
\end{table}

\subsection{Result and Analysis}\label{ssec:exp_result}

The main results of our experiments are listed in Tab.~\ref{tab:mainresult}. As mentioned in \cref{ssec:exp_setup}, we first focus on the Argument Extraction tests, where the F1 score measures the overall performance of precision and recall. In this dimension,  \OURS (T5-L) hits the highest among all generative methods. Its T5-B variant, although yields a slightly lower result, still outperforms TANL and Text2Event when they adopt pre-trained language models at the same scale. When we expand the scope to baselines of all kinds, \OURS (T5-L) ranks 2\textsuperscript{nd} among the 13 approaches benchmarked. Despite it downperforms the state-of-the-art RCEE\_ER, we argue that while the latter is a QA-based algorithm that needs templates carefully designed by experts for strong \textit{a priori}, \OURS is fully data-driven and maximally reduces human labour, which is, by all means, more accessible.

To understand the model behaviours in better detail, we additionally report the precision and recall, both of which are missing in many baseline studies. We observe that \OURS (both the T5-B and T5-L versions) achieves record-breaking recall in Argument Extraction. To be concrete, \OURS (T5-L) exceeds the previous state-of-the-art method, Text2Event (which is also a generative model based on T5-L), by a huge margin of 6.4\%. This gain is particularly valuable for commercial applications that are intolerant towards signal omissions. On the other side of the coin, we find that the precision of \OURS is relatively weak, though it is still higher than baselines such as Text2Event (T5-B). One possible cause is that, errors via if incorrectly detected event types may propagate to the downstream extraction modules (see \cref{sec:methdology}). We aim to dive deeper into this phenomenon in the upcoming research.

Although the Trigger Extraction results are less important in practice, we still investigate them for further insights. In terms of F1 score, we show that whilst \OURS does not stand out, it still yields performance that is on par with or even better than more complex baselines. As for the results of recall and precision, \OURS (both with T5-B and T5-L) again shoots the best recall among all tests approaches but fails to obtain high precision. One interesting finding is, the T5-B version of GDAP, whose scale is smaller, performs better than its T5-L counterpart in all metrics of Trigger Extraction. We will try to uncover the reasons in the future. 

\begin{table}[t]
	\small
	\centering
	\caption{
		Results of ablation studies on Argument Extraction. For reference, we duplicate the scores of \OURS in the base setup (see Tab.~\ref{tab:mainresult}), which are highlighted in colour.
	}
	\label{tab:ablationresult}
	\begin{tabular}{lcccccc}		
		\hline
		 (\%) & \textbf{P} & \textbf{R} & \textbf{F1} \\
		\hline
		\OURS (T5-B) & \cellcolor{BlueGreen!15}{47.3} & \cellcolor{BlueGreen!15}{59.1} & \cellcolor{BlueGreen!15}{52.6} \\
		\OURS (T5-L) & \cellcolor{BlueGreen!15}{48.0} & \cellcolor{BlueGreen!15}{61.6} & \cellcolor{BlueGreen!15}{54.0} \\
		\hline
		\textit{+ Golden event types} \\
		\hline
		RCEE\_ER & 69.6 & 68.4 & 69.0 \\
		\hdashline
		\OURS (T5-B) & 68.6 & 69.8 & 69.2 \\
		\OURS (T5-L) & 69.0 & 74.2 & 71.5 \\
		\hline
		\textit{- Test samples w/o events}\\
		\hline
		\OURS (T5-B) & 57.0 & 59.1 & 58.1 \\
		\OURS (T5-L) & 58.9 & 61.6 & 59.7 \\
		\hline
		\textit{- Negative sampling}\\
		\hline
		\OURS (T5-B) & 45.2 & 56.1 & 50.1 \\
		\OURS (T5-L) & 45.4 & 62.5 & 52.6 \\
		\hline
	\end{tabular}
\end{table}

We additionally conduct three ablation studies on Argument Extraction, with results exhibited in Tab.~\ref{tab:ablationresult}. To begin with, we provide golden event type annotations to RCEE\_ER and \OURS as external signals during inference. It is not surprising that the performance of tested models rises in all aspects. However, we note that contrary to the F1 score comparison in Tab.~\ref{tab:mainresult}, both the T5-B and T5-L versions of \OURS now outperform RCEE\_ER. This justifies our aforesaid assumption that the state-of-the-art RCEE\_ER does benefit a lot from manually introduced \textit{a priori}, whereas \OURS may be less precise due to errors in Event Type Detection.
 
To further demonstrate how event type errors affect model performance, from the test set we remove sentences that are not linked to any event. This adjustment lowers the chance of \OURS being \textit{misled} to predict wrong event types. As expected, the precision of \OURS instantly jumps by around 10\%. Lastly, we downgrade our proposed framework by omitting the negative sampling step. Although the overall impact on recall is not substantial, we see a precision drop for both T5-B and T5-L variants, which highlights the usefulness of our negative sampling technique. Note that removing prompts does not pose a feasible ablation setup, as the model can no longer decide whether to extract triggers or arguments. 
\section{Conclusion and Future Work}
\label{sec:con}
In this paper, we propose a novel \OURS model that attempts prompt-based learning in the Event Extraction domain for the first time. This simple method also innovatively decouples the generation of triggers and arguments, which solves the issue of target overlapping and is proved to be effective in comprehensive experiments with 11 diverse baselines. In the future, we will continue investigating our empirical observations discussed in \cref{ssec:exp_result}. Moreover, we plan to explore model weight sharing across different modules, improve the performance (especially the precision) of the \OURS framework, and transfer it to more applications.

\section{Acknowledgement}
This work is supported by the NSFC through grant (No.61872022). The Google Cloud TPU team generously provided TPU machine access for our experiments.
We would also like to express our sincerest gratitude to Guanyi Chen, Mali Jin, Yida Mu, Ruizhe Li, and the anonymous reviewers for their insightful and helpful comments.

\bibliographystyle{IEEEbib}
\bibliography{refs}

\appendix
\renewcommand\thefigure{\thesection}    
\setcounter{figure}{0}   

\section{Notation Details}

\begin{figure}[!ht]
    \centering
    \includegraphics[width=0.5\textwidth]{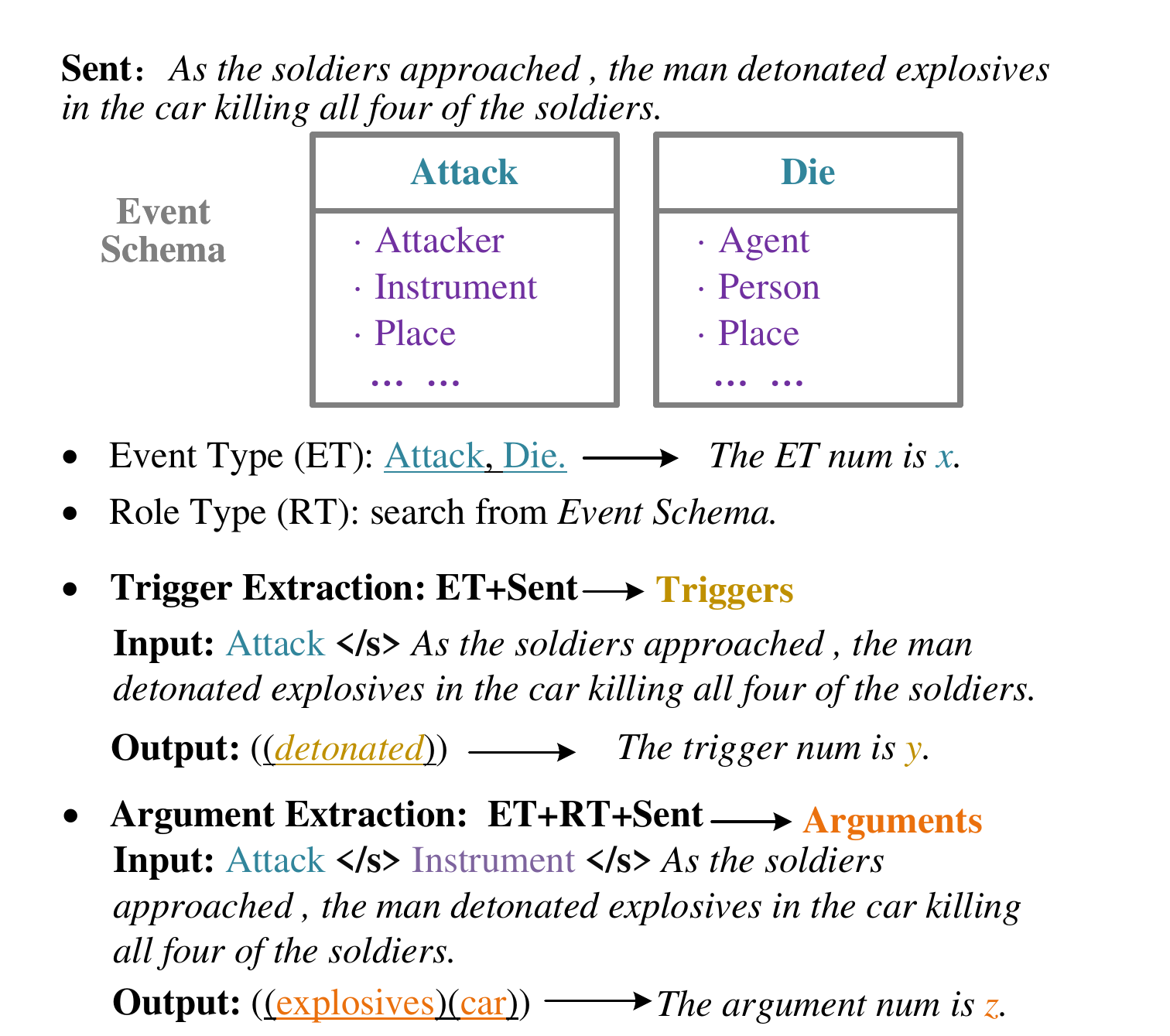}
    \caption{
        Real-world example: $x=2$, $y=1$, $z=2$.
        }
    \label{fig:xyz}
\end{figure}

\begin{enumerate}[noitemsep, topsep=0pt, partopsep=0pt,  label={$\bullet$}]
    \item \noindent\textbf{\underline{$x$}}: Number of event types in the input sentence.
    \item \noindent\textbf{\underline{$y$}}: Number of triggers corresponding to the given event type.
    \item \noindent\textbf{\underline{$z$}}: Number of obtained arguments for each type of role corresponding to the given type of event. As described in the ``Argument Extraction'' paragraph, we query the model for these arguments using prompts which starts with the concatenation of the event type, and the role type.
\end{enumerate}
See Fig.~\ref{fig:xyz} for the real-world example.


\section{Supplementary for Decoding Side}

\begin{figure}[!ht]
    \centering
    \label{fig:decoder}
    \includegraphics[width=0.5\textwidth]{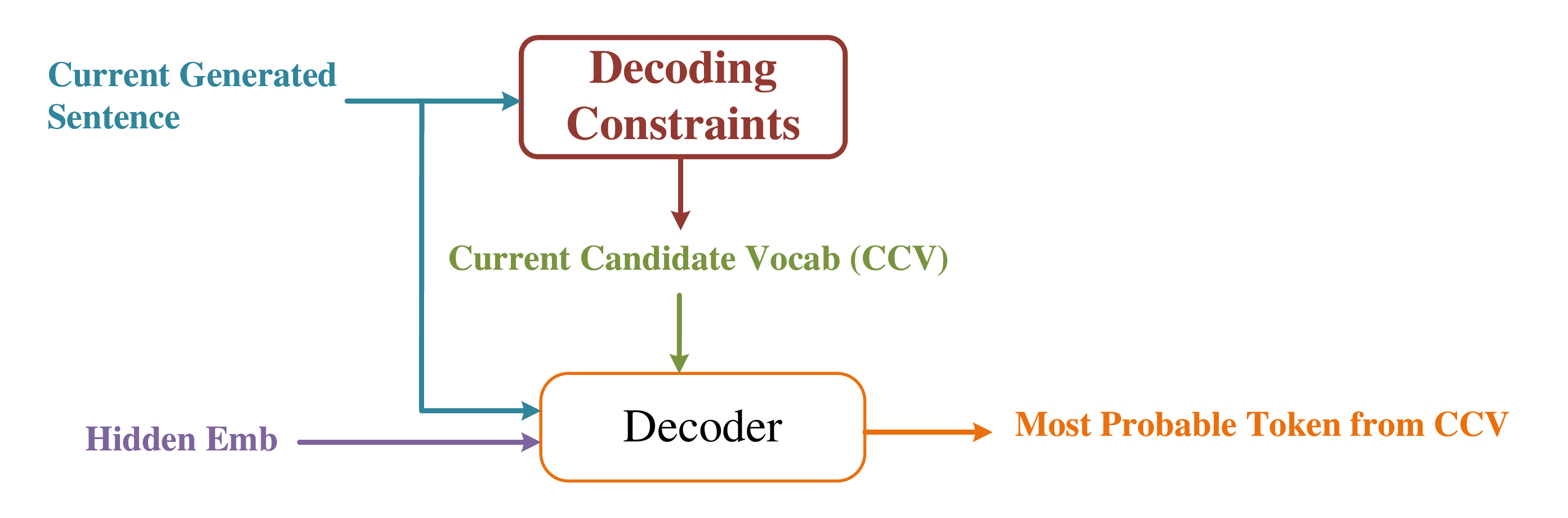}
    \caption{
        The interactions between ``Decoder'' and ``Decoding Constraints'' in Fig.~\ref{fig:overview}.
        }
\end{figure}

In the paper, we have addressed that the decoding constraints influence the token production stage and the tire-based constraint decoding algorithm is applied. That is to say, the constraints will directly limit the vocabulary of the decoder during each step, as illustrated in the Fig.~\ref{fig:decoder}.
This interaction mechanism, which is illustrated in Fig.~\ref{fig:decoder}, is not within the novelties we aim to celebrate most.

\end{document}